\pdfoutput=1

\PassOptionsToPackage{xcdraw,table}{xcolor}

\documentclass[twoside,11pt]{article}
\usepackage[final]{EMNLP2023}
\usepackage{multirow}
\usepackage{lscape}
\usepackage{xspace}
\usepackage{url}
\usepackage{lscape}
\usepackage{pdflscape,array,booktabs}
\usepackage{afterpage}
\usepackage{capt-of}
\usepackage{makecell}
\usepackage{rotating}

\usepackage{adjustbox}

\usepackage{times}
\usepackage{latexsym}
\usepackage{algorithm}
\usepackage{algpseudocode}
\usepackage{tikz}
\def\checkmark{\tikz\fill[scale=0.4](0,.35) -- (.25,0) -- (1,.7) -- (.25,.15) -- cycle;} 

\usepackage[utf8]{inputenc}
\usepackage{comment}
\usepackage{amsmath} 

\usepackage{microtype}

\usepackage{inconsolata}
\usepackage{graphicx}
\usepackage{booktabs}

\newcommand{\llamathreebase}{\textsc{Llama-3}\xspace}

\newcommand{\flores}{\textsc{FLORES-200}\xspace}
\newcommand{\mflores}{\textsc{2M-FLORES}\xspace}
\newcommand{\floresp}{\textsc{FLORES+}\xspace}

\newcommand{\bouquet}{BOUQuET\xspace}

\newcommand{\sourcebouquet}{Source-BOUQuET\xspace}
\newcommand{\fullbouquet}{Full-BOUQuET\xspace}

\title{BOUQuET \protect\includegraphics[height=1em]{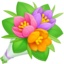}: dataset, Benchmark and Open initiative for Universal Quality Evaluation in Translation}
\author{The Omnilingual MT Team, Pierre Andrews, Mikel Artetxe, Mariano Coria Meglioli,\\ \textbf{Marta R. Costa-juss\`a, Joe Chuang, David Dale, Mark Dupenthaler, Nate Ekberg,} \\\textbf{Cynthia Gao, Daniel Licht, Jean Maillard, Alexandre Mourachko, Christophe Ropers,}\\ \textbf{Safiyyah Saleem, Eduardo Sánchez, Ioannis Tsiamas, Arina Turkatenko, }\\ \textbf{Albert Ventayol-Boada, Shireen Yates} \\ FAIR, Meta; University of London (UCL); University of the Basque Country (UPV/EHU) \\ \texttt{{costajussa}@meta.com}}

\begin{document}
\maketitle

\begin{abstract}
\bouquet is a multi-way, multicentric and multi-register/domain dataset and benchmark, and a broader collaborative initiative. This dataset is handcrafted in 8 non-English languages. Each of these source languages are representative of the most widely spoken ones and therefore they have the potential to serve as pivot languages that will enable more accurate translations. The dataset is multicentric to enforce representation of multilingual language features. In addition, the dataset goes beyond the sentence level, as it is organized in paragraphs of various lengths. Compared with related machine translation datasets, we show that \bouquet has a broader representation of domains while simplifying the translation task for non-experts. Therefore, \bouquet is specially suitable for crowd-source extension for which we are launching a call aiming at collecting a multi-way parallel corpus covering any written language. The dataset is freely available at \url{https://huggingface.co/datasets/facebook/bouquet}.
\end{abstract}

\section{Introduction}

Although multilingual large language model (LLM) evaluation benchmarks are only starting \citep{dac2023okapi}, there is a rich research history in multilingual evaluation datasets for natural language processing; e.g., \cite{sun-duh-2020-clirmatrix,malmasi-etal-2022-multiconer,yu-etal-2022-beyond}, with Machine Translation (MT) being the task with the highest investment in multilinguality \citep{kocmi-etal-2024-findings}. This is evident from the nearly 20-year history of the Conference on Machine Translation (formerly a workshop, WMT), which has established an international evaluation campaign \citep{kocmi-etal-2024-findings}. The campaign has compiled a comprehensive collection of parallel corpus evaluations covering a broad range of language pairs, domains, tasks and recently, investing in a multi-way parallel dataset expanding in languages \cite{deutsch2025wmt24expandinglanguagecoverage}. 
However, the largest multi-way parallel evaluation dataset to date was introduced with FLORES-101 \citep{goyal-etal-2022-flores}, later expanded to \flores \citep{nllb}, \floresp\footnote{\url{https://oldi.org/}} and to \mflores \citep{costajussà20242mbelebelehighlymultilingualspeech}. 

These existing datasets and benchmarks fall short due to having an English-centric focus, a narrow selection of registers, compromised quality from automated construction and mining, limited language coverage, or a static nature, in addition to being prone to contamination \cite{sainz-etal-2023-nlp}. 
Similarly, in parallel with the previous progress, there have been several initiatives that called for data annotation in a collaborative and open way, such as the translation data collection initiative \cite{singh-etal-2024-aya}. 

Recently, \citet{wu2025bitterlessonlearned2000} evaluate multilingual benchmarking and make a call for action for the need for accurate, contamination-free, challenging, practically relevant, linguistically diverse, and culturally authentic evaluations. 
This call and the urgent need of progressing in multilingual benchmarking set the stage for the introduction of a new multilingual multi-way parallel evaluation dataset and benchmark. \bouquet, which additionally combines community efforts, relies on text written from scratch (contamination-free\footnote{Note that \bouquet is free from contamination in each initial state because it is originally created and not mined. However, from the moment we open-source certain splits, \bouquet will risk to leak into training. Therefore, we keep one split hidden to avoid this.}) by native speakers in 8 different major languages (linguistically diverse). Text includes a variety of 8 practical domains (practically relevant) that represent localised knowledge (culturally diverse). \bouquet is aligned at the sentence and paragraph-level and it relies on a mixture of commissioned and openly collected human annotations to extend to any language.  

The organisation of the paper is as follows. First, the paper details how we develop the \sourcebouquet dataset (Section \ref{sec:source-bouquet}), which is the necessary stepping stone towards an open initiative. Second, we benchmark \bouquet for the 8 pivot languages plus English (Section \ref{sec:benchmark}). Finally, Section \ref{sec:openinitiative} presents how we design the open initiative itself, which aims to build the \fullbouquet dataset; i.e., \sourcebouquet translated into any written language. At the time of submission of this paper, \bouquet includes 55 multi-way parallel completed languages (Table \ref{tab:prioritylang}). 

\section{Definitions and background}

\paragraph{Definitions} Before describing the \sourcebouquet dataset's characteristics and building methodology, we define our use of some frequently encountered terms that may cover a variety of meanings.

\subparagraph{Domain.} By the term \textit{domain}, we mean different spaces in which language is produced in speech, sign, or writing (e.g., books, social media, news, Wikipedia, organization websites, official documents, direct messaging, texting). In this paper, we focus solely on the written modality.

\subparagraph{Register.} We understand the term \textit{register} as a functional variety of language that includes socio-semiotic properties, as expressed in \citet{halliday-2004}, or more simply as a ``contextual style,'' as presented in \citet[pp.79–99]{labov-1991}. In that regard, a register is a specific variety of language used to best fit a specific communicative purpose in a specific situation.


\paragraph{Background} There is a large body of work in creating datasets for MT evaluation (e.g. WMT International Evaluation Campaigns \cite{deutsch2025wmt24expandinglanguagecoverage}). However, the vast majority are limited to a few languages. we next discuss the main efforts to build massively multilingual MT benchmarks and one representation of multi-domain dataset.

\subparagraph{\floresp} \floresp \citep{maillard-etal-2024-findings} is the largest multilingual extension of \flores \cite{goyal-etal-2022-flores} and it covers the largest multi-way parallel dataset in terms of languages in 3 domains (Wikipedia, News, Travel guides). Even if \floresp has paragraph information, the translation has been done at the level of sentence without showing context to the annotators.

\subparagraph{NTREX-128} Similarly to \floresp NTREX-128 covers a multi-way parallel dataset but for 128 languages. Unlike \flores, translators had the full context of the document available when translating sentences, but the authors did not know if (or to what extent) they used this information \cite{federmann-etal-2022-ntrex}.

\subparagraph{NLLB-MD} was motivated to complement \flores in terms of domains in the context of the NLLB \cite{nllb} project. It covers chat, news and health domains in 6 languages and it includes a much larger number of sentences.


All these datasets are English-localised and English-centric, meaning that all languages have been translated from the original source English. They cover limited amount of domains (a maximum of 4) and do not differentiate among registers.

\section{Dataset: \sourcebouquet}
\label{sec:source-bouquet}

In this section, we describe the creation criteria that have been followed to design \sourcebouquet, as well as the languages it includes.

\subsection{Main characteristics}
As described in greater detail next, the \sourcebouquet dataset is mainly characterized by its non-English-centric focus, its diverse range of registers and domains (which are complementary to \flores), its manual and original composition, and its built-in dynamic extensibility.
Table \ref{tab:statistics} provides a comparison of several relevant statistics from \bouquet and the closest related datasets covered in the previous section. 

\paragraph{Non-English-centric focus.} \sourcebouquet is handcrafted by proficient speakers of Egyptian Arabic and MSA, 
French, German, Hindi, Indonesian, Mandarin Chinese, Russian, and Spanish. Each of these languages contributes the same number of sentences to the final dataset. The languages for \sourcebouquet (see Table \ref{tab:lang} in Section \ref{sec:source-bouquet-lang}) are all part of the top 20 languages in the world in terms of user population, as listed in \citet{ethnologue-27}. In addition, they are also used by a large number of non-native speakers, which makes them good candidates for what we refer to as \textit{pivot} languages; i.e., higher-resource languages that can facilitate---as source languages---the translation of datasets into lower-resource languages. English is often used as such a pivot language, since numerous people have a high degree of proficiency in English as a second language. English is not the only language in this situation, however, and is not always the best pivot language option. For example, it is much easier to find Guarani-Spanish bilingual speakers than it is to find Guarani-English bilingual speakers. What is more, cultural proximity may also make translation slightly easier.   

\paragraph{Diverse registers and domains.} Registers derive from communicative purposes and, as such, are related to domains. However, the relationship between registers and domains is not one to one. See the register and domain correspondence in Figure \ref{fig:coveredregisters} (Appendix \ref{app:registerdetails}). For example, if we take a domain such as TV news, we can identify at least 3 registers:
(1) the register used by the news anchor, which is represented by fully scripted language that is read from a teleprompter with a very specific and unnatural form of diction (e.g., hypercorrect enunciation, unnatural intonation, homogeneous pace); (2) The register produced by communication specialists (i.e., people who have been trained to be spokespersons or surrogates). The points they make have been scripted 
and rehearsed to the point of being known by heart. 
It sounds spontaneous but it is not structured like informal language; (3) the register represented in person-in-the-street segments, which is more informal and spontaneous (possibly colloquial).
This example is taken from a domain where both speech and writing are used but the situation is not significantly different in the written modality only. Language users all commonly shift between registers, which is typically referred to as \text{style-shifting}. Style-shifting (i.e., register-shifting) occurs within domains; so the domain itself is not a fool-proof way of getting a specific register. Although the norms of the domain can impose the degree of formality and of lexical specialization, it is often the register (which derives from the communicative purpose), not the domain, that determines many aspects of linguistic structure (e.g., lexical density, pronoun use, syntax, etc.).

\paragraph{Manual construction and original composition (not crawled) with accurate revisions} To develop \sourcebouquet, we set a variety of linguistic criteria that need to be covered, including both unmarked and marked structures (e.g., expected and unexpected number agreement between subject and verb). Guidelines are then shared with linguists who manually craft sentences covering examples of these linguistic criteria and compose paragraphs ranging from 3 to 6 sentences in length. These paragraphs are then manually translated across all pivot languages.

The main strategies for open collaboration are to design contribution guidelines and build an annotation tool that enables the free collection of translations in any language. 
\bouquet is shared in a repository that allows language community to easily add a new language by translating it from one of the 8 pivot languages or the English translation.
This repository contains detailed guidelines on how to do it. \bouquet's innovative approach ensures widespread language accessibility. This open collaborative initiative will enrich \bouquet with the following characteristics.

\paragraph{Language coverage extensibility}  Using both private and community-driven initiatives, we could potentially support any written language, as long as there is individual interest in contributing to multilingual advancements. 

\paragraph{Dynamic in nature} Since \bouquet includes the community, it can continuously evolve by constantly engaging it.

\begin{table*}[ht!]
\centering
\scriptsize
\begin{minipage}{\textwidth} 
\renewcommand*\footnoterule{}
\begin{tabular}{llcccccccc}
\toprule
\textsc{Dataset} &\textsc{Split} &   \textsc{$\#$Parag.}&  \textsc{$\#$Sent} & \textsc{Avg. Wrd. Parag/Sent} &  \textsc{Reg.} &  \textsc{Dom.} & \textsc{Lang.} & \textsc{Dyn.} \\
\midrule
\multirow{3}{*}{\textsc{\floresp}} & Dev &   \multirow{3}{*}{$\times$}& 997   & & &\multirow{3}{*}{Wikipedia, News, Travel guides} & \multirow{3}{*}{220} & \multirow{3}{*}{$\checkmark$} \\
& Devtest& &1,012 &  25 & $\times$\\
& Eval & & 992 &\\
\addlinespace[0.3em]
\textsc{NTREX-128} & Test & 123 & 1,997    & 389/24 & $\times$& News & 128 & $\times$ \\
\addlinespace[0.3em]
& Dev & & 6,000  \\
\textsc{NLLB-MD}& Devtest & $\times$ & 1,310 & 25 &$\times$ & Chat, News, Health & 6 & $\times$ \\ 
& Eval & & 1,500  \\
\addlinespace[0.3em]
\multirow{3}{*}{\bouquet} & Dev & 120& 504& & \multirow{3}{*}{$\checkmark$} & Fiction, Conversation, Social media & \multirow{3}{*}{55+\footnote{\scriptsize{See Appendix \ref{app:languages} for language coverage details}} }& \multirow{3}{*}{$\checkmark$}  \\
&Devtest & 200 &864& 55/15& &posts/comments, Tutorials, Website,  &\\
&Eval &144& 628 & & & Reflection pieces, Miscellaneous \\
\bottomrule
\end{tabular}%
\end{minipage}
\caption{Main statistics from MT evaluation datasets including \bouquet: number of sentences, number of paragraphs, average word per paragraph (or sentence), register information, domains, languages, dynamism. \label{tab:statistics} }
\end{table*}

\subsection{Creation criteria}
\label{sec:creationcriteria}

For the design of the creation guidelines, detailed in Appendix \ref{app:guidance}, we prepared a list of linguistic coverage requirements along with some statistical information.

\paragraph{Linguistic coverage requirements.} In order for \bouquet to be representative of various linguistic phenomena, linguistic coverage requirements are defined (as listed in Table \ref{tab:linguistic}), which are to be included in sentences that form paragraphs. 
Sentences are assigned a unique identifier that combines a unique paragraph ID number with a serial sentence number. Thus, paragraphs can be retrieved by concatenating sentences that share the same paragraph ID. 

\begin{table} [h!]
\scriptsize
\centering
\begin{tabular}{l}
\toprule
\textsc{Phenomena}  \\
\midrule
 Paragraph-like continuity\\ 
\addlinespace[0.3em]
 Variation in sentence lengths \\
 \addlinespace[0.3em]
 Dominant (unmarked) and non-dominant (marked) word orders \\ 
 \addlinespace[0.3em]
Different emphasis or topicalization \\ \addlinespace[0.3em]
 Different sentence structures (affirmation, \\interrogation, negation, subordination, coordination) \\ \addlinespace[0.3em]
 Different verb moods, tenses, and aspects \\ \addlinespace[0.3em]
 Different morphosyntactic options \\\addlinespace[0.3em]
 Different grammatical persons (1st, 2nd, 3rd, singular, plural) \\ \addlinespace[0.3em]
 Different grammatical genders \\ \addlinespace[0.3em]
 Different grammatical number agreement\\ \addlinespace[0.3em]
 Different grammatical case or forms of inflection \\\addlinespace[0.3em]
 Most frequent words used in various registers\\\addlinespace[0.3em]
 Presence of named entities, numbers, slang, and emojis \\\addlinespace[0.3em]
\bottomrule
\end{tabular}%
\caption{ \sourcebouquet Linguistic Requirements\label{tab:linguistic}}
\end{table}

\paragraph{Variety of domains.} \sourcebouquet is intended to cover 8 domains: narration (as in fiction writing), dialog, social media posts, social media comments, how-to manuals and instructions, miscellaneous website content (excluding social media or news), opinion pieces, and other miscellaneous (such as written speeches or signage). The choice of these domains optimizes for variety and popular usefulness. 

\paragraph{Variety of registers.} \sourcebouquet is built with register variety in mind, differently from \flores, which covers a few different domains but remains largely within similar registers. We characterize the registers through 3 main features (connectedness, preparedness, and social differential). Connectedness attempts to describe the type of interaction typically available in a given domain. Preparedness aims to gauge how much time is typically used to produce or edit language content. Social differential describes the relationship between the interlocutors involved in a given social situation (e.g., writer and reader, characters in a dialog, etc.). Each individual domain can present different combinations of features but  become differentiated at the level of the sentence. There are a variety of feature combinations, which are mentioned in Figure \ref{fig:coveredregisters} and defined in Appendix \ref{app:registerdetails}. 

By including new registers and domains, the new dataset is likely to be more generalizable to different contexts and applications.

\paragraph{Statistical guidance for domain representation.} In order to appropriately cover linguistic requirements and adequately represent domains, we performed a statistical analysis to understand the linguistic characteristics of each domain before creating \bouquet. In particular, our analysis covers 
most domains that we are including in \sourcebouquet by using diverse public datasets: narration (Books3, Gutenberg library \citep{gutenberg}); Social media posts (Reddit \citep{reddit}); Social media comments (Wikipedia comments\footnote{\url{https://www.kaggle.com/competitions/jigsaw-multilingual-toxic-comment-classification}}); Conversations / Dialogues (dialogsum \citep{chen-etal-2021-dialogsum-challenge}, Open Orca \citep{OpenOrca}); Tutorials/how-to articles (how-to Wikipedia-lingua \footnote{\url{https://huggingface.co/datasets/GEM/wiki_lingua}}); Website content (C4 \citep{c4}); News / Reflection pieces (CNN-DailyMail \citep{cnndaily}, XSum \citep{xsum}) and Miscellaneous (Wikipedia). Note that we collect information from public data that do not always accurately match our categories but constitute a proxy. 
For each of these domains, we have analyzed dimensionality: characters per token; tokens per sentence and sentences per paragraph; and linguistic complexity with CEFR levels\footnote{\url{https://www.coe.int/en/web/common-european-framework-reference-languages/}}.

\begin{figure}[h!]
 \includegraphics[width=0.28\textwidth]{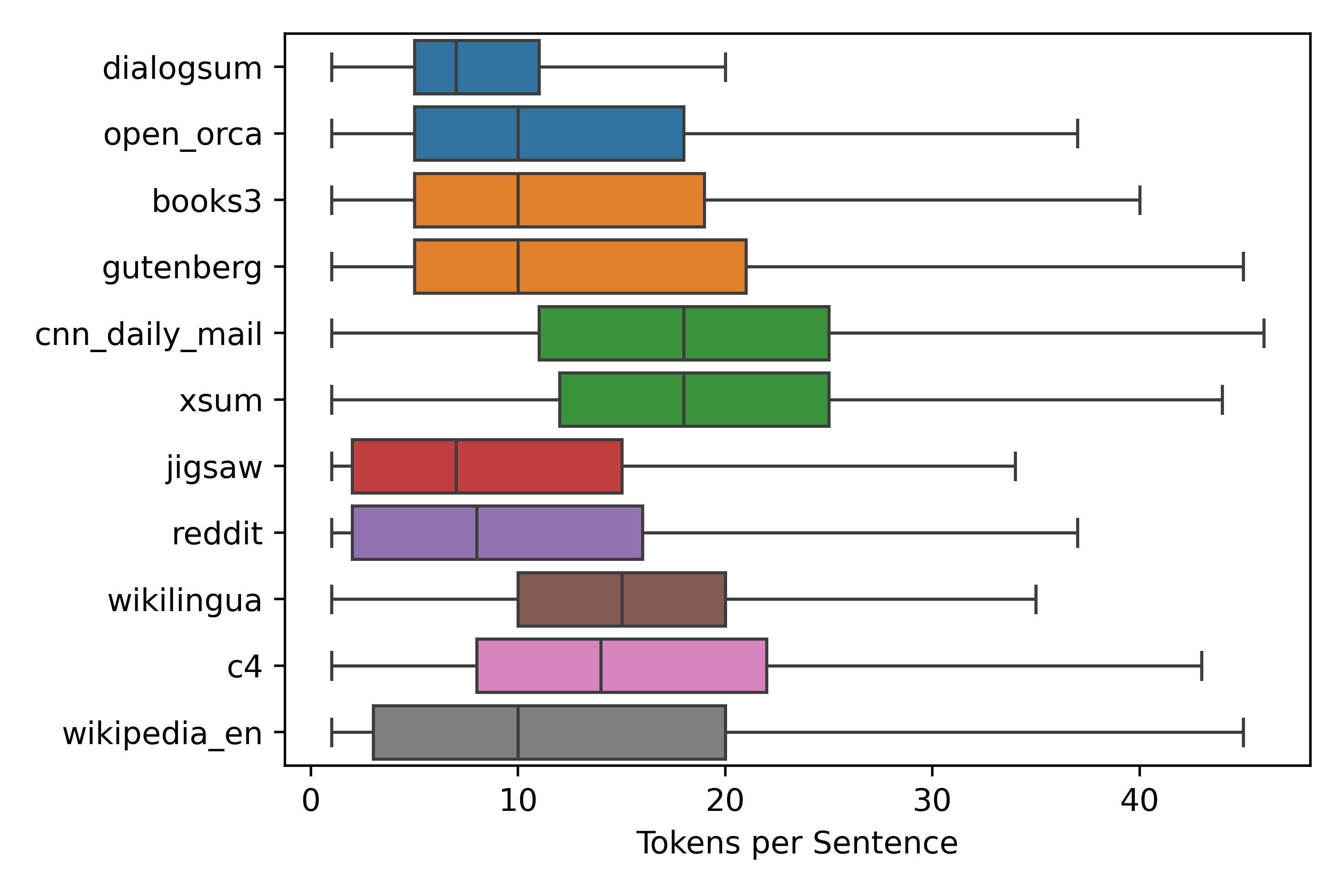} 
  \includegraphics[width=0.28\textwidth]{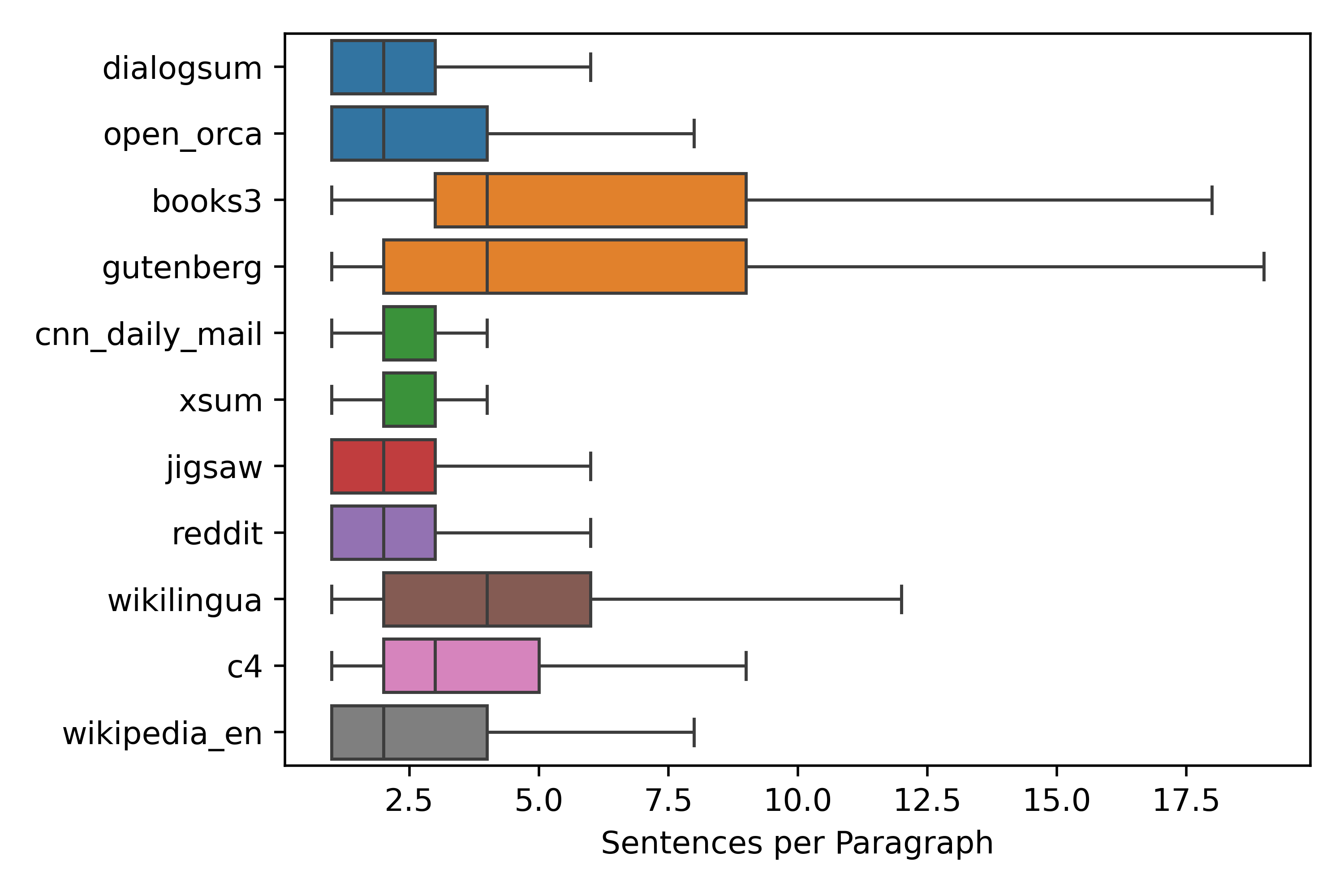}
   \includegraphics[width=0.28\textwidth]{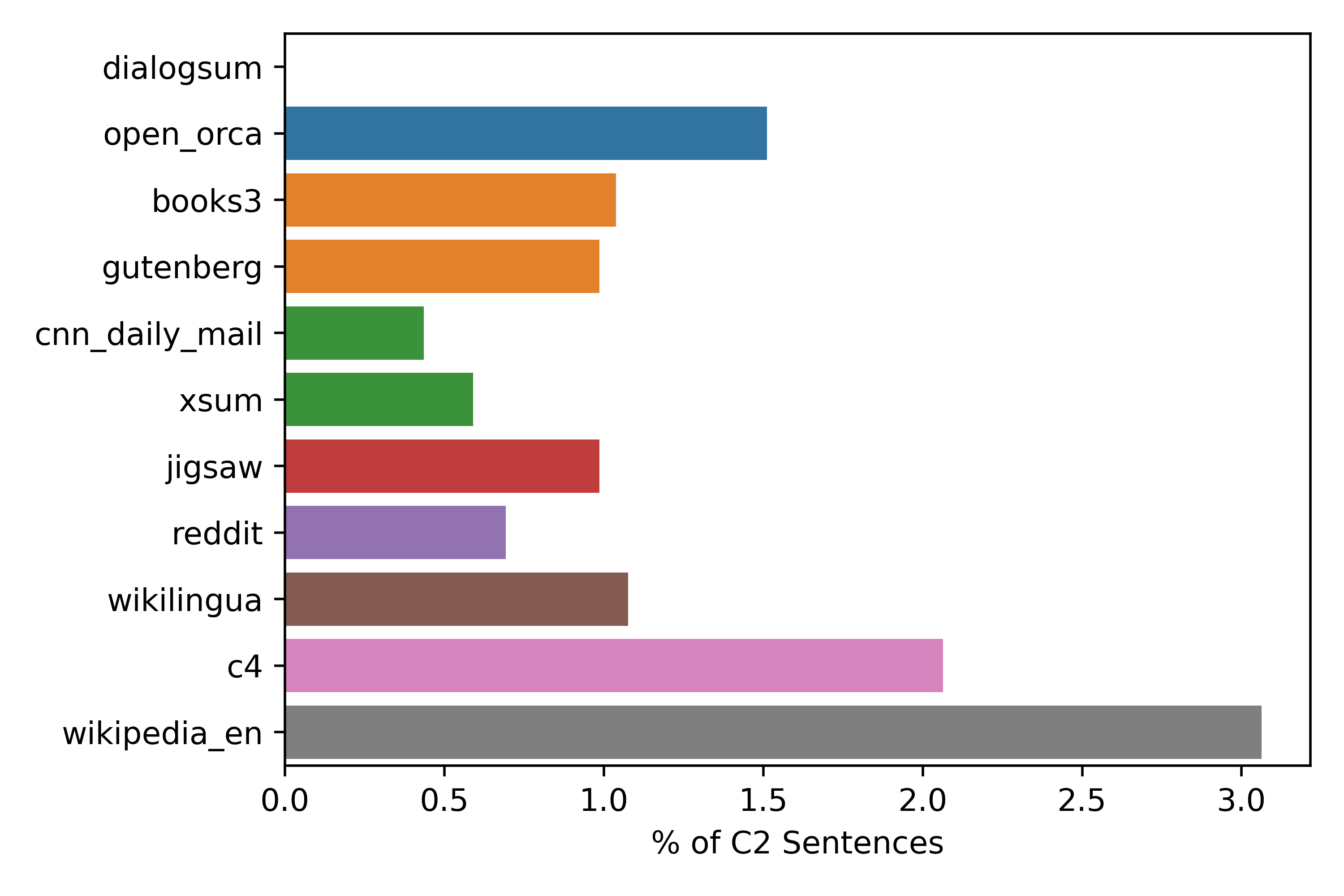} 
   \includegraphics[width=0.10\textwidth]{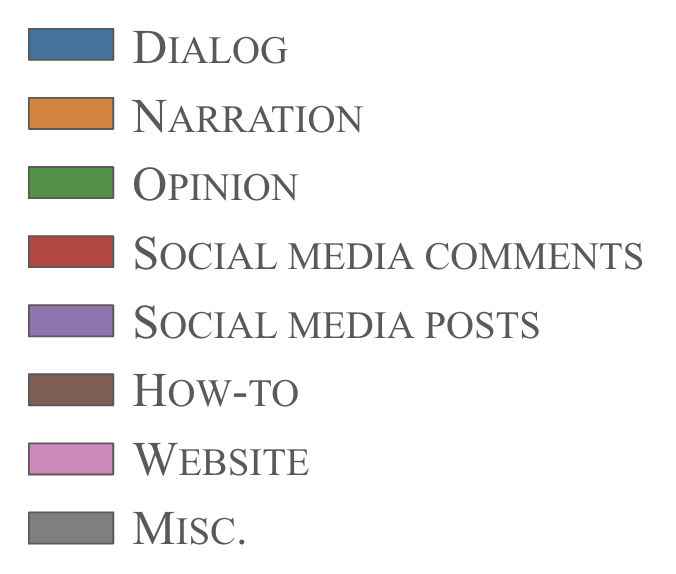} 
  \caption{ (Top) Tokens per sentence and (Middle) sentences per paragraph (Bottom) CEFR per dataset representative of \bouquet domains.
\label{fig:chrtokens}}
\end{figure}

Regarding tokens per sentence (Figure \ref{fig:chrtokens} left), we can see correlations between different domains, and clear differences in length, especially in dialogs which tend to be much shorter. Regarding sentences per paragraph (Figure \ref{fig:chrtokens} middle), we can find a correlation between different datasets representing the same domain, where fiction writing paragraphs tend to be much longer (averaging 5 but reaching up to 20~), dialogs and news articles are much shorter (barely reaching 3-4 sentences in a paragraph), and the rest of the categories are somewhere in between (normally staying between 1-5 but reaching up to 10 in some cases).

To guide \bouquet creators on the linguistic complexity required for each domain, we have assessed complexity using the distribution of CEFR levels as a proxy. This includes the \% of C2 scores at the sentence level for each dataset. These scores were labeled by a SONAR-based model \citep{duquenne2023sonarsentencelevelmultimodallanguageagnostic} trained on CEFR-SP data for CEFR Text Classification, which significantly outperformed \llamathreebase \citep{touvron2023llama}. See right side of Figure \ref{fig:chrtokens}.
Wikipedia seems to be the only dataset with a more considerable share of C2 sentences, with some others like dialogues having no samples scored as such.

\paragraph{Annotations and Quality Checks} Each entry of \sourcebouquet includes the source text (in one of the 8 pivot languages of Table \ref{tab:lang}) and its translation into English, domain information and contextual information for better translation accuracy. 
To double-check that \sourcebouquet does not contain repeated sentences, we explored the similarity across English sentences. For each English sentence, we computed SONAR embeddings \cite{duquenne2023sonarsentencelevelmultimodallanguageagnostic} and we computed the cosine distance on the vectors. There were only 14 sentences with a cosine distance below 0.3. These sentences are reported in Appendix \ref{app:examples}.

\subsection{Languages} 
\label{sec:source-bouquet-lang}

As mentioned earlier, \bouquet aims to be multicentric and localized , in contrast to most existing datasets that are English-centric. The motivation is mainly to be representative of linguistic phenomena. To this effect, it is created in 8 non-English languages (Table \ref{tab:lang}). Each language contributes with a similar number of sentences along with their English equivalents given by the sentence creators themselves. 

\subsection{Multi-way extension to \sourcebouquet languages}
\label{sec:multiway}

\paragraph{Details} \sourcebouquet creators composed 250 sentences for each of the 8 pivot languages plus the corresponding English translation. The remaining 1,750 sentences for each pivot language are translated from English. The final \sourcebouquet is composed of 2,000 sentences in 9 languages (8 pivot languages plus their translations into English). 

\paragraph{Quality checks} Since multi-way parallel data is created from English, we manually checked that translations did not lose the linguistic information when translating from English.
While translating \bouquet, we had to make sure that the contextual information which was applicable to the whole paragraph was taken into consideration by the translators. To ensure this, we used a number of following QA strategies reported in Appendix \ref{app:qualitychecks}.

\paragraph{Additional contextual information}  The multi-centric nature of \bouquet is also a reminder that English is not morphologically rich (e.g., it doesn't mark grammatical gender agreement between nouns, adjectives, and verbs) and displays relatively little information about formality in its written form (e.g., it uses only one second-person singular pronoun, regardless of who is addressing whom). As such, English isn't an ideal source language for translation purposes unless translators can be provided with additional contextual information. The \bouquet dataset includes such additional information; for example, the grammatical gender of the first and second person (when this isn't obvious) or the linguistic markedness of some words or phrases (e.g., literary or archaic verb tenses, use of slang, infrequently used level of formality).


\begin{table} [h!]
\centering
\scriptsize
\begin{tabular}{lllllllllll}
\toprule
\textsc{ISO} & \textsc{ISO} & \textsc{Language} & \textsc{Family} & \textsc{Subgroup1} \\
\textsc{6393} & \textsc{15924} & \\
\midrule
\addlinespace[0.3em]
arb	& Arab & \multirow{1}{*}{Modern }	& \multirow{1}{*}{Afro-Asiatic}	& \multirow{1}{*}{West Semitic}\\
&&Standard Arabic \\
cmn	&Hans &	Mandarin 	& Sino-Tibetan & Sinitic	\\
& &	 Chinese	& \\
\addlinespace[0.3em]
deu & Latn & German	& Indo-European	& West Germanic \\
\addlinespace[0.3em]
fra	& Latn &	French	& Indo-European	& Italic \\
\addlinespace[0.3em]
hin	& Deva &	Hindi &	Indo-European&	Indo-Aryan	\\
    \addlinespace[0.3em]
ind	& Latn & Indonesian &	Austronesian &	Malayic \\
\addlinespace[0.3em]
rus	& Cyrl &	Russian&	Indo-European	& Balto-Slavic	\\
\addlinespace[0.3em]
spa	& Latn &	Spanish	& Indo-European	& Italic \\
\bottomrule
\end{tabular}%

\caption{ \sourcebouquet Languages\label{tab:lang}}
\end{table}

\subsection{Overall Statistics}  

\begin{figure}[h!]
\centering
   \includegraphics[width=0.5\textwidth]{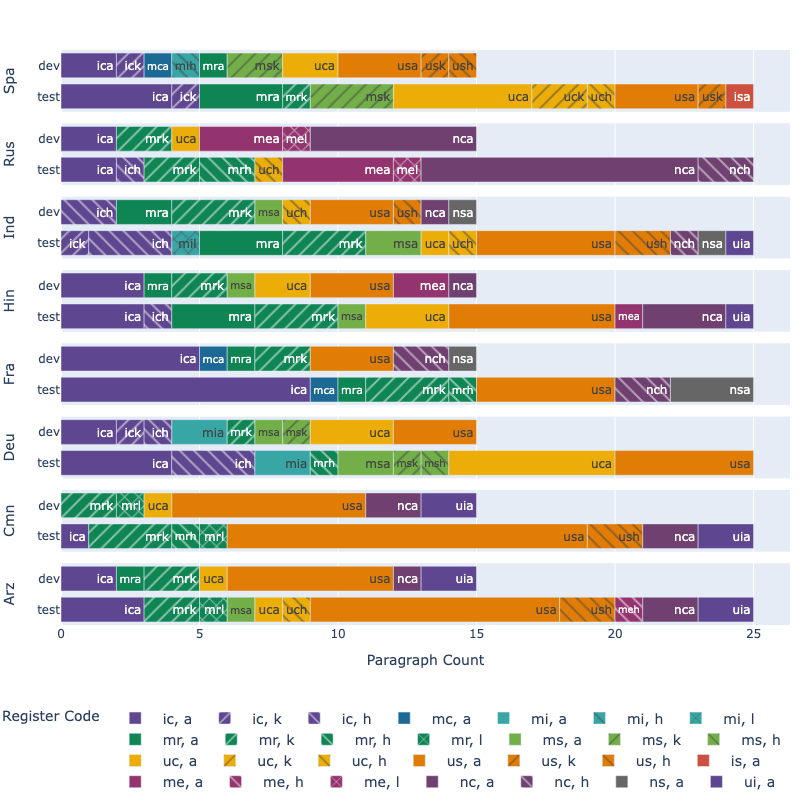} 
  \includegraphics[width=0.5\textwidth]{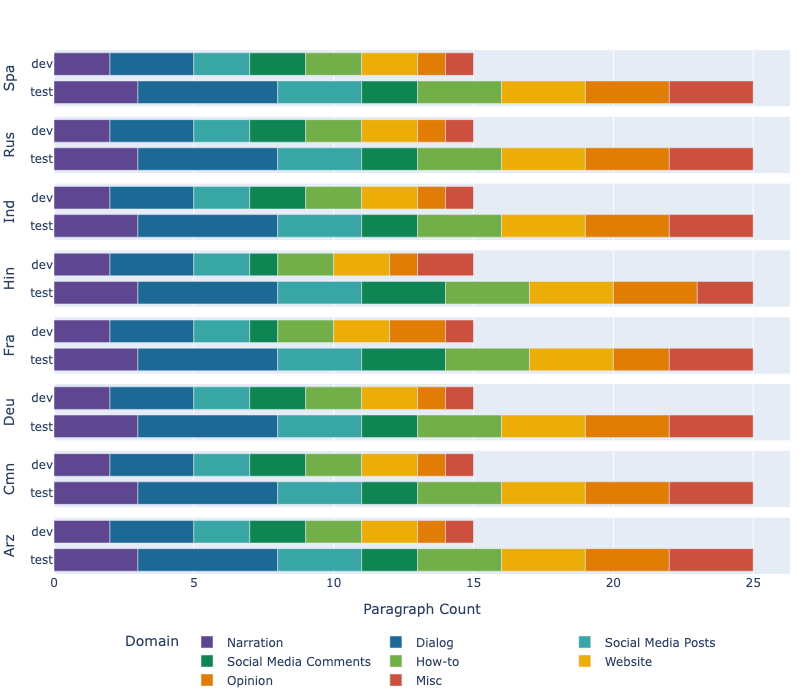} 
  \caption{\label{fig:splits} Registers (top) and domains (bottom) representations in development and test partitions.}
\end{figure}

{In total, \bouquet contains 2,000 sentences. These sentences are split by making a stratified paragraph-level selection among source languages and domains into development, test and evaluation sets.  Initially, the evaluation set (632/144 sentences/paragraphs) is intended to be kept hidden. Figure \ref{fig:splits} shows the representation of registers (top) and domains (bottom) in the non-hidden splits. Labels for each of the combinations of register options are created by concatenating the lowercase letters used as unique identifiers (see details of these register options in the Appendix \ref{app:registerdetails}).  For example, a register characterized as impersonal (in connectedness), composed (in preparedness), and equal-assumed (in social differential) is labeled \texttt{ica}.}

The results in the following section are presented with the test split of 864/200 sentences/paragraphs.


\section{Benchmark}
\label{sec:benchmark}

We benchmark \bouquet in two dimensions: domain representation and machine translation. The former quantifies how representative \bouquet is of public datasets of multiple domains compared to other evaluation datasets. The latter addresses how several MT systems are ranked with \bouquet compared to other evaluation datasets. 

\begin{figure}[h!]
\centering
  \includegraphics[width=0.5\textwidth]{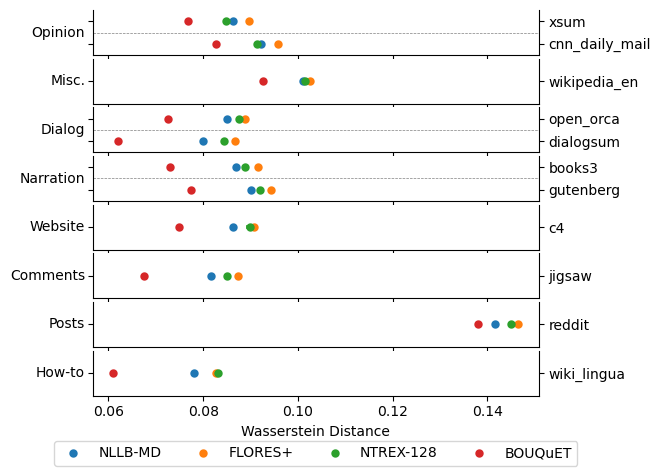}
  \caption{Wasserstein Distance (WD) for each domain and dataset. Lower WD indicate better representation of the domain. \label{fig:wd-distribution}}
\end{figure}

{\paragraph{Domain representation} The performance of the model in a new or unseen dataset depends on the similarity between the dataset that was used to fit the model and the new dataset.  We compare the domain coverage of \bouquet with that of \floresp, NTREX-128 and NLLB-MD. To do this comparison, we take a random sample of 2,000 sentences (which seems to be a sufficiently large sample of the embedding space for score stability) from each of the  domain datasets from Figure \ref{fig:chrtokens}; as well as 2000 from each alternative dataset \floresp, NTREX-128, NLLB-MD,
and \bouquet. We create vector representations of each sentence in previous datasets with SONAR~\citep{duquenne2023sonarsentencelevelmultimodallanguageagnostic}. 
From SONAR vectors, we do a PCA-dimensionality reduction, fitted upon the combined multi-domain set, see Figure \ref{fig:domrep} (Appendix \ref{app:domrepdetails}). Public domains from Figure \ref{fig:chrtokens} are represented in grey; alternatives evaluation datasets are represented in blue and \bouquet is represented in red. Figure \ref{fig:domrep}, from top to down, compares \bouquet against \flores, NTREX-128, NLLB-MD, 
respectively. 
We qualitatively observe that \bouquet covers a wider range of domains.
Additionally, to quantify this coverage, we measure the overlap between each dataset with each of the domains using the Wasserstein distance (implemented with the POT library\footnote{\url{https://pythonot.github.io/}}). The Wasserstein Distance (WD), also known as the Earth Mover’s Distance (EMD), is a metric that measures the "effort" required to transform one probability distribution into another. Lower results indicate a higher similarity between clusters. This distance is run on the full 1024-dimensional SONAR embedding vectors, without applying any kind of dimensionality reduction (PCA).
Some domain sets and evaluation sets were several orders of magnitude larger than each other.  Sampling all down to the same size (2,000) makes the metric computable in a reasonable amount of time and removes any sensitivity to class imbalance in the distribution distance metric. 
Figure \ref{fig:wd-distribution} shows that the lowest consistent results are obtained for all domains with \bouquet.}



\paragraph{Machine Translation} To help the reader understand why the dataset is useful, we present preliminary results to demonstrate its use for its intended purpose: MT benchmarking. We evaluate 14 translation systems: 
\llamathreebase (Llama3.1-8B, Llama3.2-3B, Llama3.3-70B)\citep{touvron2023llama}, Tower (TowerInstruct-7B-v0.2) \citep{rei-etal-2024-tower}, Aya (Aya101-13B, Aya-Expanse-8B) \citep{aya}, Babel (Babel-9B-Chat) \citep{zhao2025babelopenmultilinguallarge}, Cohere (CohereLabs-command-r7b-12-2024), Eurollm (EuroLLM-9B-Instruct) \citep{martins2024eurollmmultilinguallanguagemodels}, MADLAD (MADLAD-3B-MT and MADLAD-10B-MT) \citep{kudugunta2023madlad400multilingualdocumentlevellarge}, Mistral (Mistral-7B-Instruct-v0.3)\footnote{\url{https://docs.mistral.ai/getting-started/models/models_overview/}}, Qwen (Qwen2.5-7B-Instruct) \citep{bai2023qwentechnicalreport}
and NLLB (NLLB-3.3B) \citep{nllb}. We select the models as ones with open weights, focusing primarily on moderate sizes (about 10B) and variety of architectures. Following the official evaluation metrics of WMT 2024 \citep{kocmi-etal-2024-findings}, we use two automatic metrics: CometKiwi (CometKiwi-da-xl, range 0-1 and $\uparrow$ better, COM) \citep{chimoto-bassett-2022-comet} and MetricX (MetricX-24-hybrid-xl-v2p6, range 0-25 and $\downarrow$ better, MetX) \citep{juraska-etal-2024-metricx}. We include in the benchmarking datasets that cover \sourcebouquet languages (\floresp and NTREX-128).

Table \ref{tab:results} shows that \bouquet scores consistently higher than other datasets on average, suggesting that \bouquet is easier to translate. This is an advantage for the open initiative, since the complexity of current MT test sets makes it harder to ask the community to participate in translations as it requires a high-level of expertise.  

Rankings across models and datasets is not preserved, which hints that all datasets may be posing different challenges to the models. Rankings is computed as counting when a system is similar in the same position according to CometKiwi. This ranking and Pearson correlation on the CometKiwi is dissimilar for datasets evaluated at the sentence-level, with \bouquet being the most different. This difference is enlarged when evaluating at the paragraph-level where number of swaps increases and, coherently, Pearson correlation decreases, meaning that datasets pose different challenges to models. We need to further investigate which linguistic challenges \bouquet is adding. However, best two systems are consistent across datasets and level of evaluation (sentence and pargraphs) being those the largest model (Llama3.3-70B) and Aya-e-8B. 

NLLB-3.3B has a higher variation between being evaluated at the sentence or paragraph-level, which makes sense since it is the only one trained with sentence-level data.

\begin{table*}[ht!]
\centering
\scriptsize
\begin{tabular}{l|cccccc|cccc}
\toprule
 Model&  \multicolumn{2}{c}{\textsc{Bouquet}} &  \multicolumn{2}{c}{\textsc{Flores}} &  \multicolumn{2}{c}{\textsc{Ntrex}}  &  \multicolumn{2}{|c}{\textsc{BouquetP}}  &  \multicolumn{2}{c}{\textsc{NtrexP}}\\
 & {\tiny \textsc{COM}} & {\tiny \textsc{MetX}}  & {\tiny \textsc{COM}} & {\tiny \textsc{MetX}}   & {\tiny \textsc{COM}} & {\tiny \textsc{MetX}}    & {\tiny \textsc{COM}} & {\tiny \textsc{MetX}}   & {\tiny \textsc{COM}} & {\tiny \textsc{MetX}}    \\
\midrule
\hline 
NLLB-3B & 0.68 & 2.1  & 0.66 & 2.56 & 0.65 & 2.97 & 0.59 & 3.71  & 0.29 & 14.1 \\
	aya101-13B & 0.67 & 2.02 & 0.63 & 2.65 & 0.63 & 3.14 & 0.58 & 3.29 & 0.24 & 13.17\\
	aya-e-8B & 0.69 & \textbf{1.75} & 0.65 & 2.9 & \textbf{0.67} & \textbf{2.45} & 0.64 & \textbf{2.42} & 0.34 & \textbf{8.7}\\
	babel-9B & 0.67 & 2.33& 0.65 & 2.66 & 0.63 & 3.36& 0.61 & 3.4 & 0.32 & 10.39\\
	cohere-7B & 0.67 & 2.15 & 0.65 & 2.89 & 0.64 & 3.01& 0.61 & 3.2  & 0.32 & 9.61 \\
	eurollm-9B & 0.67 & 2.33& 0.65 & 2.89  & 0.61 & 3.64& 0.61 & 3.64 & 0.31 & 10.08 \\
	madlad-10B & 0.63 & 2.74 & 0.64 & 2.72 & 0.63 & 3.35& 0.41 & 6.76& 0.15 & 15.99 \\
	madlad-3B & 0.63 & 2.85 & 0.63 & 2.94 & 0.61 & 3.67 & 0.37 & 6.71& 0.49 & 5.29\\
	mistral-7B & 0.54 & 4.29 & 0.51 & 5.69 & 0.49 & 6.12 & 0.49 & 6.64& 0.24 & 10.96  \\
	qwen-7B & 0.59 & 3.25& 0.6 & 3.75  & 0.59 & 4.21& 0.57 & 4.5 & 0.52 & 4.93\\
	Llama3.1-8B & 0.66 & 2.36  & 0.64 & 2.82& 0.63 & 3.27& 0.6 & 3.33  & 0.32 & 10.17 \\
	Llama3.2-3B & 0.59 & 3.59 & 0.57 & 4.34 & 0.55 & 4.89& 0.52 & 5.52&  0.27 & 12.67\\
	Llama3.3-70B & \textbf{0.7} & 1.85& \textbf{0.68} & \textbf{2.21} & \textbf{0.67} & 2.59 & \textbf{0.63} & 2.72 &  \textbf{0.35} & 9.76 \\
	Tower-7B & 0.58 & 3.69 & 0.56 & 4.19& 0.56 & 4.35& 0.49 & 5.65 & 0.28 & 12.22 \\ \hline \hline
    
           &  \multicolumn{2}{c}{\textsc{Bouquet-Flores}} & \multicolumn{2}{c}{\textsc{Flores-Ntrex}} & \multicolumn{2}{c}{\textsc{Ntrex-Bouquet}} &\multicolumn{4}{c}{\textsc{BouquetP-NtrexP}}  \\ \hline
           Swaps & \multicolumn{2}{c}{3} & \multicolumn{2}{c}{4} & \multicolumn{2}{c}{7} & \multicolumn{4}{c}{11} \\ 
           Pearson Cor. & \multicolumn{2}{c}{0.95} & \multicolumn{2}{c}{0.99} & \multicolumn{2}{c}{0.95} & \multicolumn{4}{c}{0.92}  \\ 
        \hline
\end{tabular}
        \caption{Averaged Results XX-to-XX 9 \sourcebouquet (8 pivot plus English) languages for \bouquet, \floresp, NTREX-128 at the level of sentence 2 columns on the left and at the level of paragraph 3 columns on the right. Number of ranking swaps (a system not being in the same position according to CometKiwi) from each dataset compared to the other two (in similar sentence or paragraph-level) and Pearson correlation indicate that while datasets report similar results at sentence-level, being \bouquet the most different, it is not the case for paragraph-level where the ranking of systems varies by a larger amount. \label{tab:results}}
\end{table*}

\begin{figure}[h!]
\centering
 \includegraphics[width=0.45\textwidth]{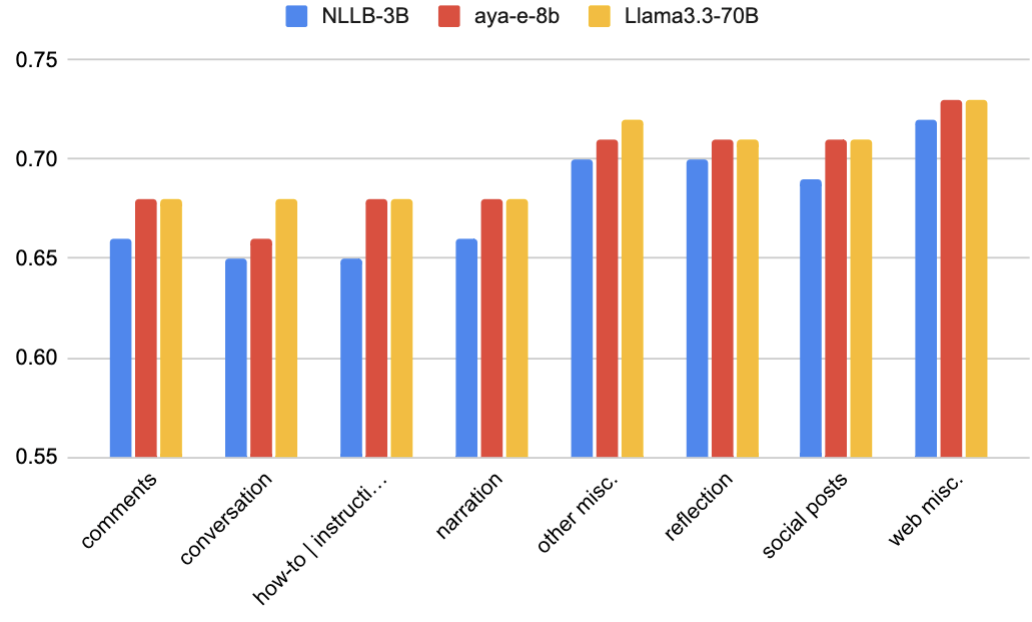}
  \caption{ Best performing models and their results in each of the \bouquet domains 
\label{fig:domres}.}
\end{figure}

Figure \ref{fig:domres} shows results of the 3 best systems averaged across language directions, evaluated at the sentence-level, per domains. Worse performing domains are comments, conversations, how-to and narration. Best performing domains are web and other miscellaneous, reflection and social posts. Appendix \ref{app:domrepdetails} reports more detailed results on \bouquet per language and domains.

\section{Beyond commissioning translations: Open initiative}
\label{sec:openinitiative}

 \sourcebouquet is intended to be translated into any written language. For this, we have commissioned an initial set of priority languages covering a variety of high and low-resource languages representing different geographical regions, linguistic families and scripts. See the list of languages currently covered by \bouquet in Appendix \ref{app:languages}. 
 
 However, it would be challenging to achieve our language coverage target to any language. This ambition can only be achieved with the support of the community. For this, we have organized an open collaborative effort which involves language communities that are interested in contributing to this effort. 
 
The purpose of this open initiative is to collect translations from \sourcebouquet.  To collect these annotations, we have set a tool to collect annotations.
Together with setting \sourcebouquet in this tool, we use the annotation guidelines from Section \ref{sec:multiway} which very much resemble those from \flores \citep{nllb} and which are available in the 9 \bouquet languages. One of the advantages is that annotators can choose the source language from among one of the \sourcebouquet languages. These languages have been chosen to cover a wide range of speakers, facilitating the task of annotation instead of depending on English bilingual speakers. 
This open initiative is available in \url{https://bouquet.metademolab.com/}. 

\section{Conclusions and Next Steps}
\label{sec:conclusions}

In this paper, we have presented the \sourcebouquet dataset and the attached open initiative. We have shown consistent gains in domain diversity in two different metrics while keeping complexity lower than its competitors. The latter is particularly relevant to simplify the translation for non-experts that may join the open initiative. We also provide MT results for the 8 languages in which \sourcebouquet has been created. Although \bouquet is currently totally completed for 55 languages (see list \ref{tab:prioritylang}), this number is only a fraction of the language coverage ambition that we are pursuing by launching the open initiative for community efforts. Please join us in making Universal Quality Evaluation in Translation available in any language.

Beyond increasing in number of languages, \bouquet is actively evolving, and we are currently working on designing quality control for each of the contributions and adding new languages to the incremental releases of \bouquet and extending the benchmarking by further showing the capabilities of \bouquet, e.g. increasing the evaluation of linguistic signals over its alternatives.

\section*{Limitations and Ethical Considerations}

The \bouquet dataset is still limited in the number of languages and translations.The benchmarking is quite complete (4 datasets comparison, 14 models and 2 metrics) but it can also be extended in several axes (linguistic analysis). However, the entire purpose of this work is to describe the dataset and open-initiative, while providing a minimal benchmarking. Authors expect the community to extend the benchmarking by further using this dataset for further exploration. Creators and commissioned translation's annotators are paid a fair rate. 

\bibliographystyle{acl_natbib}
\bibliography{anthology,custom,bib-bouquet}

\appendix

\section{Specific guidance for paragraph and sentence creation}
\label{app:guidance}

\subsection{Overview}
The Bouquet-source dataset comprises 250 unique sentences in each of its source languages. This means that each linguist created (i.e., wrote from scratch, did not copy; see Section 2.4 above) 250 original sentences. These sentences were requested to be:
\begin{itemize}
\item Organized in logically structured paragraphs (see the Paragraphs section below)
\item Representative of the linguistic structures and features most frequently used in specific domains (see the Domains section below)
\item Representative of the most common register of language used in similar situations (see the Registers section below)
\item Accompanied by a gold-standard (i.e., best in class) human translation into English.
\end{itemize}

\subsection{Paragraphs}
The linguist received a template in the form of a spreadsheet, in which paragraph structures were designed and laid out. The template specified the exact number of paragraphs and the exact number of sentences for each of the paragraphs. Each paragraph was given a unique paragraph ID (e.g., P01, P02, P15). Each sentence within each paragraph was also given a serial, non-unique ID (e.g., S1, S2, S3). 

\subsection{Domains}
The template was divided into 8 domains:
\begin{enumerate}
\item How-to, written tutorials or instructions
\item Conversations (dialogues)
\item Narration (creative writing that doesn't include dialogues)
\item Social media posts
\item Social media comments (reactive)
\item Other web content
\item Reflective piece
\item Miscellaneous (address to a nation, disaster response, etc.)
\end{enumerate}
The creators had to produce the set number of sentences for each of the domains; the structure of the template (domain / paragraph / sentence) could not be changed.

\subsection{Language Register Information}
When creating sentences, the creators had to make sure that the register of language being used was representative of the most expected and appropriate register for the situation. When several registers were possible, the creators were asked to use discretion when selecting a register, while making sure that the chosen register was among the most expected and appropriate. To help them make a determination, we defined 3 main functional areas of language register:
\begin{itemize}
\item Connectedness: What type of connection do language users who initiate the text have with other language users?
\item Preparedness: How much time do language users who initiate the text had or took to prepare the text?
\item Social differential: What is the relative social status of the language users who initiate the text towards other language users?
\end{itemize}

\subsection{Linguistic Features}
One of the main reasons for dividing the dataset into sections that correspond to domains is to attempt to cover as many registers and aspects of language as possible. For example, we know that:
\begin{itemize}
\item Some pro-drop languages may drop the subject pronouns more often in some situations than in others.
\item Some case-marking languages may use some cases in specific situations but avoid them in others.
\item In English, lexical density increases when the level of formality increases.
\item Some languages use a specific past verb tense in storytelling, which stands out from other past verb tenses used in casual conversations or other situations.
\item Some languages use specific verb moods in some situations but avoid them in others.
\end{itemize}

\subsection{Violating Content}
While creating sentences, the creators were asked to avoid inserting violating content. Violating content is language that can fall under one (or more) of the below categories:
\begin{itemize}
\item Toxicity
\item Illegal activities
\item Stereotypes and biases
\end{itemize}

\subsection{Step-by-Step Description of Tasks}
Please refer to Table \ref{tab:stepbystep} for the step-by-step description of the tasks.

\begin{table*}
\scriptsize
\begin{tabular}{|l|l|}
\hline
\textbf{Column A: Lang-ID} & This column should have the same 3-lowercase-letter code representing the source language of the sentences \\
&being created followed by an underscore character ( \_) and a 4-letter code representing the script. \\
\hline
\textbf{Column B: Domain} & This is 1 of the 8 domains represented in the dataset (see Section 3.1). \\
\hline
\textbf{Column C: Subdomain} & Please insert your description of the subdomain or topic. \\
\hline
\textbf{Column D: P-ID} & This is the unique code identifying a paragraph (e.g., P01, P02, \ldots, P58). \\
\hline
\textbf{Column E: S-ID} & This is the non-unique code identifying the sequential place of the sentence within a paragraph. \\
\hline
\textbf{Column F: Sentence} & In this cell, please type a sentence you created. \\
\hline
\textbf{Column G: Translation into English} & After entering a sentence in your language in Column F, please provide a gold-standard human translation \\& in this cell. \\
\hline
\textbf{Column H: S-Nchars} & This represents a count of the number of characters in the sentence. \\
\hline
\textbf{Column I: S Comment\_src\_lang} & To help other linguists expand this dataset by translating your sentences into their own languages,\\& please add any comments that bring more context about the sentence. \\
\hline
\textbf{Column J: S Comment\_English} & Please provide an English translation of the comment your inserted in Column I. \\
\hline
\textbf{Column K: Linguistic features} & Please list the register- or domain-specific linguistic features you tried to showcase in the sentence. \\
\hline
\textbf{Column L: Connectedness} & Please use any of the options best describing the register area of Correctedness. \\
\hline
\textbf{Column M: Preparedness} & Please use 1 of the options best describing the register area of Preparedness. \\
\hline
\textbf{Column N: Social differential} & Please use any of the options best describing the register area of Social differential. \\
\hline
\textbf{Column O: Formality} & Please indicate the level of formality best characterizing the sentence. \\
\hline
\textbf{Column P: Relationship} & Please insert the intended relationship between the language users involved in the situation. \\
\hline
\textbf{Column Q: Idea origin} & Please insert the name of the media type or platform that inspired the sentence. \\
\hline
\textbf{Column R: P Comment\_src\_lang} & To help other linguists expand this dataset by translating your sentences into their own languages, \\& please add any comments that bring more context about the entire paragraph. \\
\hline
\textbf{Column S: P Comment\_English} & Please provide a translation into English for the comment you inserted in Column R. \\
\hline
\textbf{Column T: P-Nchars} & This represents a count of the number of characters in the current paragraph. \\
\hline
\textbf{Column U: Creator\_Translator-ID} & Please insert your ID here, if it isn't pre-populated. \\
\hline
\end{tabular}
\caption{\label{tab:stepbystep} Step-by-step guidance.}
\end{table*}

\subsection{Additional Guidance on Domain-Specific Content}

Dialogues, especially those inserted in long creative writing (such as novels), often include the name of the speaker or a cue mark (e.g., — ), and sometimes quotation marks. When creating sentences for conversations, the creators were let free to invent names for speakers or to label speaker turns (e.g., A, B); they were also asked to place the names or speaker reference in markup tags, similarly to this: \textless Name:\textgreater or \textless A:\textgreater.

\textbf{Emojis}: As there are emojis frequently in some social media and messaging domains, some representation was also expected from the creators. However, the creators were asked to keep this representation very limited, as there are no real agreed ways to translate them across hundreds of languages.

\textbf{Social media comments}: The creators were told that they could keep the structure of those comments flat, and that including tags was not absolutely necessary, though it was permitted (even expected).

\textbf{Disfluencies in informal conversations}: Disfluencies were permitted provided they were representative of conversations and they could be translated (i.e., there is some consensus on how to write them in the language — ah, oh, um).

\section{Registers Details}
\label{app:registerdetails}

\begin{figure*}[h!]
\centering
 \includegraphics[width=1\textwidth]{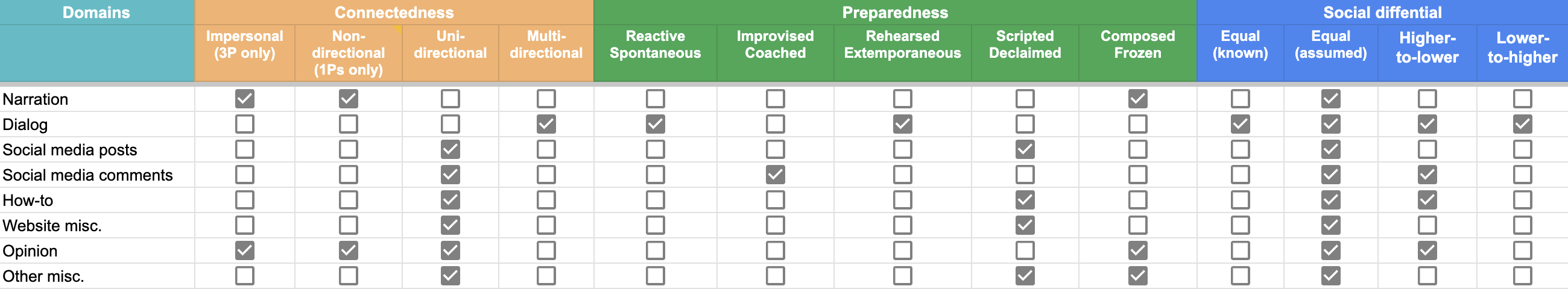} 
  \caption{Register functional areas and breakdowns within each functional areas and their representations across domains.
\label{fig:coveredregisters}}
\end{figure*}

We provide non-exclusive options for each of the 3 functional areas that characterize registers described in Section \ref{sec:creationcriteria} and mentioned in Figure \ref{fig:coveredregisters}. By non-exclusive, we mean that a domain may be characterized by more than one option. The functional area / option breakdown can be described as follows (the bold lowercase letters in square brackets represent a unique identifier for each option):
\subsection*{Connectedness}
\begin{itemize}
\item Impersonal [\textbf{i}]: For example a text written for the purpose of giving definitions or explanations with no specific readership in mind; typically written in the third person only (e.g., a contract).
\item Non-directional [\textbf{n}]: A text written with a readership in mind but that doesn't address the readership specifically (e.g., an author recounting a story)
\item Uni-directional [\textbf{u}]: A text addressing a readership who either cannot respond or is asked to refrain from responding at a given time (e.g., the transcription of a presentation, such as a TED Talk)
\item Multi-directional [\textbf{m}]: A text addressing a readership who can respond (e.g., SMS, DM) or representing the transcription of a dialogue involving 2 or more language users.
\end{itemize}
\subsection*{Preparedness}
\begin{itemize}
\item Reactive (spontaneous) [\textbf{r}]: The production is immediate either because it needs to be or because the user wants it to be
\item Improvised (coached) [\textbf{i}]: The production appears spontaneous but takes place after a period of general training or coaching (e.g., spokespeople who answer questions live but have had time to prepare and choose vocabulary to use or to avoid)
\item Rehearsed (extemporaneous) [\textbf{e}]: The production is live but its overall structure has been carefully crafted and rehearsed (e.g., transcriptions of 20-minute presentations or speeches that aren't fully scripted and given from notes).
\item Scripted (declaimed) [\textbf{s}]: The production may or may not be live and has been fully scripted (e.g., transcriptions of speeches used in teleprompters)
\item Composed (frozen) [\textbf{c}]: The production is completely offline, and goes through iterations of reviewing and editing (e.g., the text of a novel).
\end{itemize}
\subsection*{Social differential}
\begin{itemize}
\item Equal (known) [\textbf{k}]: The readership or addressees are known to be peers; this can include a very informal or colloquial attitude
\item Equal (assumed) [\textbf{a}]: The readership or addressees are not known but assumed to be peers; this can include a casual or informal attitude but likely excludes a very colloquial one
\item Higher-to-lower [\textbf{h}]: The readership or addressees are considered to be at a lower social level than the producer (e.g., the producer is arrogant or assumes a position of higher authority)
\item Lower-to-higher [\textbf{l}]: The readership or addressees are considered to be at a higher social level than the producer (e.g., the producer wants to express deference, respect, or admiration)
\end{itemize}

\section{Quality Checks Details in Multi-way extension}
\label{app:qualitychecks}

In order to make sure that the \bouquet contextual information was taken into accound while translating \bouquet, we used the following QA strategies:

\begin{enumerate}
    \item Checking the correct co-referencing. The Bouquet dataset is a representation of natural language, and the usage of personal and possessive pronouns as a substitute for the nouns is a typical occurrence. If the internal co-referencing in the paragraph is broken (the wrong pronoun is used or the noun is repeated where the noun should be), it indicates that the paragraph was treated as a collection of sentences not linked to each other, rather than a paragraph of text.

\item Checking the lexical consistency. We made sure to check that vocabulary used to translate word denoting objects or events is appropriate in tone, style and register and is used consistently throughout each paragraph. For example, when checking, we found out that translations from Indonesian into Russian did not keep consistency for “potato fritters” (“perkedel kentang”), using three different ways to translate it in P-292. We later applied the necessary corrections.

\item Checking the grammatical consistency. Since the Bouquet dataset contains examples of different domains, we needed to check whether the verb tenses and syntax were appropriate for a given domain and used consistently throughout each paragraph. For example, when checking translated into German paragraphs which imitate fiction narration, we made sure that German Pratäritum tense is used appropriately, not Perfekt.

\item Checking the special symbols such as emojis and numbers. 
\end{enumerate}

\section{Priority Languages}
\label{app:languages}

Table \ref{tab:prioritylang} shows the languages in which \bouquet exists at the time of submission of this paper (May 2025).

\begin{table*} [h!]
\centering
\scriptsize
\begin{tabular}{llllllllllll}
\toprule
\textsc{ISO 639-3} & \textsc{ISO 15924} & \textsc{Language} & \textsc{Family} & \textsc{Subgroup} & Class \\
\midrule
\addlinespace[0.3em]
arz	& \multirow{2}{*}{Arab}& Egyptian Arabic & \multirow{2}{*}{Afro-Asiatic} & \multirow{2}{*}{Central Semitic} & \multirow{2}{*}{Pivot}\\
(+ arb	&  & +Modern Stan. Arabic)\\
arz & Latn & Romanized Egyptian Arabic & Afro-Asiatic & Semitic & P1-HR \\
\addlinespace[0.3em]
aar & Latn & Afar & Afro-Asiatic & Cushitic & P1-LR \\
\addlinespace[0.3em]
agr & Latn & Aguaruna & Chicham & -- & P1-LR\\
\addlinespace[0.3em]
ami & Latn & Amis & Austronesian & East Formosan & P1-LR\\
\addlinespace[0.3em]
ben & Beng & Bengali & Indo-European & Indo-Aryan & P1-HR \\
\addlinespace[0.3em]
cmn	& Hans & Mandarin Chinese & Sino-Tibetan & Sinitic & Pivot	\\
\addlinespace[0.3em]
ces & Latn & Czech & Indo-European & Balto-Slavic & P1-HR \\
\addlinespace[0.3em]
crk & Cans & Plains Cree & Algic & Algonquian & P1-LR\\
\addlinespace[0.3em]
deu & Latn & German	& Indo-European	& West Germanic & Pivot \\
\addlinespace[0.3em]
dje & Arab, Latn & Zarma & Songhay & Eastern Songhay & P1-LR\\
\addlinespace[0.3em]
ell & Grek & Modern Greek & Indo-European& Hellenic & P1-HR \\
\addlinespace[0.3em]
fra	& Latn & French	& Indo-European	& Italic & Pivot \\
\addlinespace[0.3em]
gaz & Latn & West Central Oromo & Afro-Asiatic & Cushitic & P1-LR\\
\addlinespace[0.3em]
gil & Latn & Gilbertese & Austronesian & Micronesian & P1-LR\\
\addlinespace[0.3em]
guc & Latn & Wayuu & Arawakan & Caribbean Arawakan & P1-LR\\
\addlinespace[0.3em]
hin	& Deva & Hindi & Indo-European & Indo-Aryan	& Pivot \\
    \addlinespace[0.3em]
    hin & Latn & Romanized Hindi & Indo-European& Indo-Aryan & P1-HR \\
\addlinespace[0.3em]
hrv & Latn & Croatian & Indo-European & Balto-Slavic & P1-HR \\
\addlinespace[0.3em]
hun & Latn & Hungarian & Uralic & Hungaric & P1-HR \\
\addlinespace[0.3em]
ind	& Latn & Indonesian & Austronesian & Malayic & Pivot \\
\addlinespace[0.3em]
ita & Latn & Italian & Indo-European & Italic & P1-HR \\
\addlinespace[0.3em]
jav & Latn & Javanese & Austronesian & Javanesic & P1-HR \\
\addlinespace[0.3em]
jpn & Jpan & Japanese & Japonic & & P1-HR \\
\addlinespace[0.3em]
kaa & Cyrl & Karakalpak & Turkic & Kipchak & P1-LR\\
\addlinespace[0.3em]
kal & Latn & Kalaallisut & Eskimo-Aleut & Eskimo & P1-LR\\
\addlinespace[0.3em]
khm & Khmr & Central Khmer & Austroasiatic & Mon-Khmer & P1-HR \\
\addlinespace[0.3em]
kor & Kore & Korean & Korean&   Koreanic& P1-HR \\
\addlinespace[0.3em]
kru & Deva & Kurukh & Dravidian & North Dravidian & P1-LR\\
\addlinespace[0.3em]
lij & Latn & Ligurian & Indo-European & Italic & P1-LR\\
\addlinespace[0.3em]
lin & Latn & Kinshasa Lingala & Atlantic-Congo & Central West. Bantu & P1-LR\\
\addlinespace[0.3em]
mya & Mymr & Burmese & Sino-Tibetan & Burmo-Qiangic & P1-LR\\
\addlinespace[0.3em]
nld & Latn & Standard Dutch & Indo-European & West Germanic & P1-HR \\
\addlinespace[0.3em]
pes & Arab & Western Persian & Indo-European& Iranian & P1-HR \\
\addlinespace[0.3em]
pol & Latn & Polish & Indo-European & Balto-Slavic & P1-HR \\
\addlinespace[0.3em]
rus	& Cyrl & Russian & Indo-European & Balto-Slavic & Pivot	\\
\addlinespace[0.3em]
ron & Latn & Romanian & Indo-European & Italic & P1-HR \\
\addlinespace[0.3em]
sba & Latn & Ngambay & Central Sudanic & Sara-Bongo-Bagirmi & P1-LR\\
\addlinespace[0.3em]
spa	& Latn & Spanish & Indo-European	& Italic & Pivot \\
\addlinespace[0.3em]
por & Latn &  Portuguese (Brazilian) & Indo-European & Italic & P1-HR \\
\addlinespace[0.3em]
swe & Latn & Swedish & Indo-European & North Germanic & P1-HR \\
\addlinespace[0.3em]
swh & Latn & Coastal Swahili & Atlantic-Congo & N.E. Coastal Bantu & P1-HR \\
\addlinespace[0.3em]
tha & Thai & Thai & Tai-Kadai & Southwestern Tai & P1-HR \\
\addlinespace[0.3em]
tir & Ethi & Tigrinya & Afro-Asiatic & Semitic & P1-LR\\
\addlinespace[0.3em]
tgl & Latn & Tagalog & Austronesian & Greater Central Philippine & P1-HR \\
\addlinespace[0.3em]
tur & Latn & Turkish & Turkic & Oghuz & P1-HR \\
\addlinespace[0.3em]
ukr & Cyrl & Ukrainian & Indo-European & Balto-Slavic & P1-HR \\
\addlinespace[0.3em]
urd & Arab & Urdu & Indo-European & Indo-Aryan & P1-HR \\
\addlinespace[0.3em]
vie & Latn & Vietnamese & Austroasiatic & Vietic & P1-HR \\
\addlinespace[0.3em]
yor & Latn & Yoruba & Atlantic-Congo & Defoid & P1-LR\\
\addlinespace[0.3em]
zlm & \multirow{2}{*}{Latn} & Colloquial Malay & \multirow{2}{*}{Austronesian}& \multirow{2}{*}{Malayic}& \multirow{2}{*}{P1-HR} \\
+zsm &   & + Standard Malay \\
\addlinespace[0.3em]
\bottomrule
\end{tabular}%

\caption{ \sourcebouquet Languages (Pivot) and Priority languages (P) both high-resource (HR) and low-resource (LR) \label{tab:prioritylang} included in \bouquet at the time of submission. Note that these languages have been commissioned, we do not include updates in annotations collected from the open-initiative, which we will include in later versions of the paper.}
\end{table*}

\section{Dataset Examples}
\label{app:examples}


Table \ref{tab:cd} reports the sentences with highest similarity score computed with cosine distance of SONAR vectors across all 2,000 \sourcebouquet English sentences.

\begin{table*} [h!]
\centering
\scriptsize
\begin{tabular}{llllp{0.15\textwidth}llp{0.15\textwidth}l}
\toprule
cosinedist	& lang-A &	lang-B	&Domain-A	&Text-A	&UNIQID-A	&DomainB	&Text-B	&UNIQID-B \\ \midrule
0.19&	ind &	arz &	conversation&		What time do we meet?	& P304-S4	& conversation&	when will we meet?	&P017-S1\\
	0.20 &	fra&	cmn &	web misc.&	About us	&P220-S1	&web misc.	&	About our team	&P098-S1\\
 0.22	& rus&	deu &	conversation	&$<$B:$>$ Which one?&	P363-S2	&conversation&	$<$B:$>$ When and where?&	P134-S2\\
0.24 &	rus&	fra &	conversation	& $<$B:$>$ Nah, I am sick&	P360-S2 &	conversation	&	$<$B:$>$You're sick?	&P185-S4\\
	0.25 &	rus&	fra&	conversation&		$<$A:$>$ You know what I mean!&	P362-S4	&conversation	&	$<$A:$>$Did you hear?	&P183-S1\\
	0.28&	fra &	arz &	web misc.	&	Send us your résumé and motivation letter at the below address.&	P215-S6&	web misc.	&Please sendyour CV with letters of recommendation to this email address&	P043-S5\\
	0.28&	spa	&rus&	comments	&	$<$B:$>$ WHAT IS THIS???&	P443-S2	&conversation	&	$<$B:$>$ What do you mean?	&P362-S2\\
	0.29 &	rus&	fra &	conversation	& $<$A:$>$ Get well soon&	P360-S3	&conversation&		$<$A:$>$Not doing very well.&	P185-S3\\
0.29 & rus &	fra &	conversation	&	$<$B:$>$ What do you mean?	&P362-S2	&conversation&	$<$A:$>$Did you hear?	&P183-S1\\
	0.29 &	rus&	fra&conversation&	$<$B:$>$ What do you mean?	&P362-S2&	conversation&	$<$B:$>$You're sick?&	P185-S4\\
	0.29 &	rus&	fra &	conversation&$<$B:$>$ Nothing is working for me.	&P366-S2	&conversation&		$<$A:$>$Not doing very well.	&P185-S3\\
\bottomrule
\end{tabular}%
\caption{ \sourcebouquet sentences with closest similarity score (cosine distance lower than 0.3)  \label{tab:cd}}
\end{table*}

Table \ref{tab:example} shows complete entries examples of the \sourcebouquet dataset.

\begin{table*} [h!]
\centering
\scriptsize
\begin{tabular}{llp{0.08\textwidth}llp{0.15\textwidth}p{0.15\textwidth}p{0.11\textwidth}l}
\toprule
LangID	& Domain	& Subdomain	& PID	& SID	& Sentence	& English & Linguistic label&	Reg.	\\
\midrule
  spa\_Latn &	conversation	&text message chain	&P417	&S1	&$<$Guillermo$:>$ Habéis cenado ya?	&$<$Guillermo$:>$ Have you had dinner already?	&	word:named-entity	& mrk \\
spa\_Latn &	conversation&	text message chain	&P417	&S2	&$<$Jaime$:>$ No, estábamos pensando en salir ahora, te apuntas?&	$<$Jaime$:>$ No, we were thinking about going out now. Are you in?	& 	word:named-entity	& mrk\\
spa\_Latn & conversation	&text message chain	&P417	&S3	&$<$Guillermo$:>$ Sí, me estoy muriendo de hambre.&	$<$Guillermo$:>$ Yes, I'm starving.	& 	word:named-entity, miscellaneous:collocation 	&mrk\\
spa\_Latn & conversation	&text message chain	&P417	&S4	&$<$Jaime$:>$ Guai, salimos en cinco, te esperamos en la parada del metro.	&$<$Jaime$:>$ Cool, we're leaving in five, we'll wait for you at the metro station. &	word:named-entity, word:slang	&mrk\\
spa\_Latn&	conversation&	text message chain	&P417	&S5	&$<$Guillermo$:>$ Perfecto, me cambio y salgo.	&$<$Guillermo$:>$ Perfect, I'll change and head out.	&	word:named-entity&	mrk\\
\midrule
fra\_Latn &	social posts&	Integrity&	P204&	S1&	Choses que j'aurais aimé savoir plus tôt&	Things I wish I had known earlier&	sentence:fragment&	usa\\
fra\_Latn&	social posts&	Integrity	&P204	&S2	&Si tu ne prends pas de décision pour toi-même, d'autres les prendront pour toi.	&If you don't make decisions for yourself, others will take them for you.	&	word:impersonal-pronoun&	usa\\
fra\_Latn&	social posts&	Integrity	&P204	&S3	&Quand on te submerge de généralités, demande plusieurs exemples spécifiques.	&When you are getting submerged in generalities, request several specific examples.	&word:impersonal-pronoun&	usa\\
\midrule
ind\_Latn	&narration&	Folklore / Fable	&P310	&S1	&Pada suatu masa, hiduplah sepasang suami istri di sebuah pedesaan.&	Once upon a time, there lived a husband and wife in a village.&		Third person, impersonal, narration&	ica\\
ind\_Latn	&narration&	Folklore / Fable&	P310	&S2	&Mereka belum juga dikarunia anak setelah sekian lama menikah.&	They have not yet been blessed with children after being married for so long.	&Third person, impersonal, narration&	ica\\
ind\_Latn&	narration	&Folklore / Fable&	P310	&S3	&Keduanya bermimpi bahwa mereka harus menanam timun, jika mereka ingin memiliki anak. &	Both of them dreamed that they had to plan cucumbers, if they wanted to have a child. &	Third person, impersonal, narration	&ica\\
ind\_Latn&	narration	&Folklore / Fable&	P310&	S4	&Kemudian ditanamlah timun-timun itu.&	Then they planted the cucumbers.&	Third person, impersonal, narration&	ica\\
\bottomrule
\end{tabular}%
\caption{ \bouquet examples including main fields  \label{tab:example}}
\end{table*}

\section{Domain representation details}
\label{app:domrepdetails}

\begin{figure*}[h!]
\centering
 \includegraphics[width=0.32\textwidth]{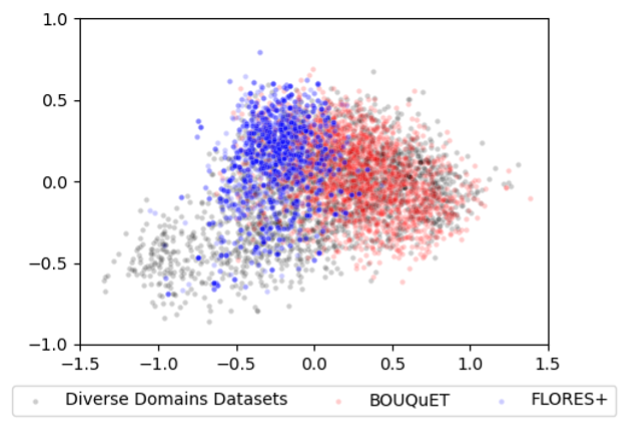}
    \includegraphics[width=0.32\textwidth]{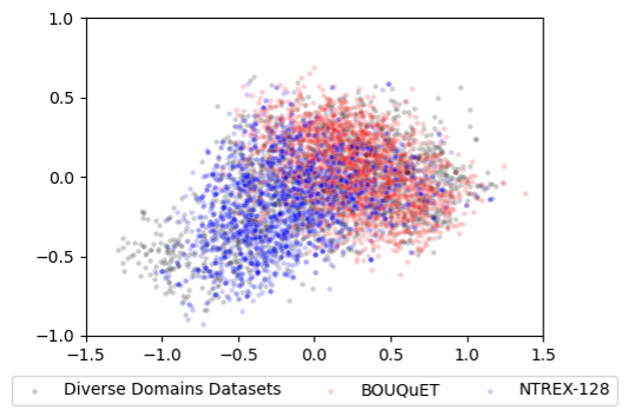}
  \includegraphics[width=0.32\textwidth]{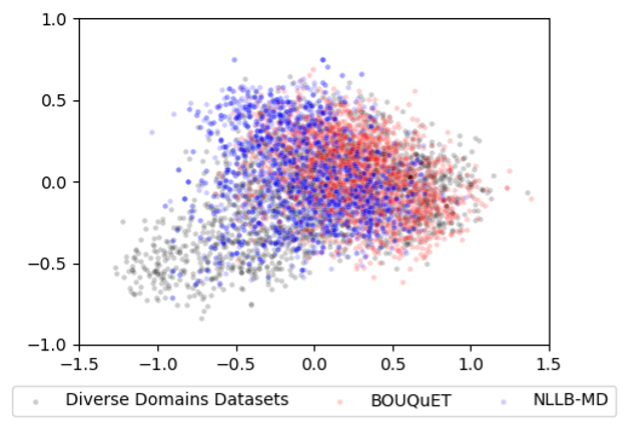}

  \caption{ Domain representation and overlap across \floresp (left), NTREX-128 (middle), NLLB-MD (right)(in blue) with diverse domains datasets (in grey) and \bouquet (in red). 
\label{fig:domrep}}
\end{figure*}

Figures \ref{fig:domrep} shows the domain representation and overlap across datasets.

\section{Detailed results}

\begin{table*}[htp]
    \centering
    \tiny
    \setlength{\tabcolsep}{3pt} 
    \resizebox{\textwidth}{!}{ 
    \begin{tabular}{ l  c  c  c  c  c  c  c  c  c  c  c  c  c  c  c  c  c  c }
        \hline
         Src-lang &  \multicolumn{2}{c}{arz-Arab} &  \multicolumn{2}{c}{cmn-Hans} &  \multicolumn{2}{c}{deu-Latn} &  \multicolumn{2}{c}{eng-Latn} &  \multicolumn{2}{c}{fra-Latn}  &  \multicolumn{2}{c}{hin-Deva} &  \multicolumn{2}{c}{ind-Latn} &  \multicolumn{2}{c}{rus-Cyrl} &  \multicolumn{2}{c}{spa-Latn}  \\
        \hline
         & \textsc{com} & \textsc{metx} & \textsc{com} & \textsc{metx} & \textsc{com} & \textsc{metx} & \textsc{com} & \textsc{metx} & \textsc{com} & \textsc{metx} & \textsc{com} & \textsc{metx} & \textsc{com} & \textsc{metx} & \textsc{com} & \textsc{metx} & \textsc{com} & \textsc{metx} \\ \hline \hline
	nllb-3B & 0.59 & 2.31 & 0.66 & 1.39 & 0.72 & 2.39 & 0.75 & 1.86 & 0.69 & 2.11 & 0.61 & 2.33 & 0.69 & 2.03 & 0.67 & 2.33 & 0.72 & 2.13 \\
	aya101-13B & 0.59 & 2.23 & 0.65 & 1.32 & 0.71 & 2.31 & 0.75 & 1.76 & 0.67 & 2.07 & 0.6 & 2.32 & 0.69 & 1.88 & 0.67 & 2.26 & 0.71 & 1.99 \\
	aya-e-8B & 0.6 & 2.04 & 0.68 & \textbf{1.09} & 0.73 & \textbf{2.06} & \textbf{0.77} & \textbf{1.47} & 0.7 & \textbf{1.77} & 0.62 & \textbf{1.94} & 0.7 & \textbf{1.69} & 0.69 & \textbf{1.94} & 0.73 & \textbf{1.76} \\
	babel-9B & 0.57 & 2.78 & 0.66 & 1.61 & 0.72 & 2.58 & 0.75 & 1.92 & 0.68 & 2.47 & 0.59 & 2.54 & 0.69 & 2.16 & 0.67 & 2.47 & 0.71 & 2.46 \\
	cohere-r7B & 0.58 & 2.38 & 0.66 & 1.28 & 0.72 & 2.45 & 0.75 & 1.87 & 0.66 & 2.52 & 0.61 & 2.18 & 0.7 & 1.86 & 0.66 & 2.65 & 0.71 & 2.16 \\
	eurollm-9B & 0.58 & 2.55 & 0.66 & 1.59 & 0.71 & 2.68 & 0.75 & 2.04 & 0.68 & 2.38 & 0.61 & 2.41 & 0.66 & 2.37 & 0.67 & 2.58 & 0.71 & 2.36 \\
	madlad-10B & 0.52 & 3.71 & 0.62 & 1.8 & 0.66 & 3.23 & 0.73 & 2.0 & 0.64 & 2.61 & 0.58 & 2.75 & 0.62 & 3.06 & 0.63 & 3.08 & 0.69 & 2.39 \\
	madlad-3B & 0.51 & 3.93 & 0.62 & 1.76 & 0.63 & 3.51 & 0.72 & 2.26 & 0.63 & 2.82 & 0.59 & 2.63 & 0.61 & 3.25 & 0.63 & 3.1 & 0.7 & 2.4 \\
	mistral-7B & 0.44 & 5.32 & 0.54 & 3.53 & 0.59 & 4.52 & 0.6 & 3.83 & 0.55 & 4.2 & 0.46 & 4.95 & 0.59 & 3.81 & 0.56 & 4.31 & 0.59 & 4.11 \\
	qwen-7B & 0.5 & 3.83 & 0.58 & 2.55 & 0.63 & 3.49 & 0.68 & 2.79 & 0.6 & 3.25 & 0.52 & 3.45 & 0.61 & 3.16 & 0.59 & 3.47 & 0.62 & 3.29 \\
Llama-3.1-8B & 0.56 & 2.77 & 0.64 & 1.58 & 0.7 & 2.63 & 0.75 & 1.88 & 0.66 & 2.46 & 0.58 & 2.77 & 0.68 & 2.24 & 0.66 & 2.52 & 0.7 & 2.38 \\
	Llama3.2-3B & 0.46 & 4.88 & 0.58 & 2.56 & 0.63 & 3.86 & 0.67 & 3.02 & 0.58 & 3.77 & 0.53 & 3.78 & 0.61 & 3.35 & 0.59 & 3.46 & 0.63 & 3.62 \\
	Llama3.3-70B & \textbf{0.61} & \textbf{1.96} & \textbf{0.68} & 1.18 & \textbf{0.74} & 2.14 & \textbf{0.77} & 1.62 & \textbf{0.71} & 1.94 & \textbf{0.62} & 2.1 & \textbf{0.71} & 1.73 & \textbf{0.69} & 2.05 & \textbf{0.74} & 1.91 \\
	Tower-7B & 0.4 & 6.07 & 0.62 & 1.5 & 0.63 & 3.23 & 0.66 & 3.48 & 0.56 & 4.21 & 0.49 & 4.42 & 0.62 & 3.26 & 0.6 & 3.61 & 0.64 & 3.46 \\
\hline
\hline
       Trg-lang &  \multicolumn{2}{c}{arz-Arab}  & \multicolumn{2}{c}{cmn-Hans} & \multicolumn{2}{c}{deu-Latn}& \multicolumn{2}{c}{eng-Latn} &  \multicolumn{2}{c}{fra-Latn} &  \multicolumn{2}{c}{hin-Deva}  &  \multicolumn{2}{c}{ind-Latn} &  \multicolumn{2}{c}{rus-Cyrl} &  \multicolumn{2}{c}{spa-Latn} \\
       \hline
        & \textsc{com} & \textsc{metx} & \textsc{com} & \textsc{metx} & \textsc{com} & \textsc{metx} & \textsc{com} & \textsc{metx} & \textsc{com} & \textsc{metx} & \textsc{com} & \textsc{metx} & \textsc{com} & \textsc{metx} & \textsc{com} & \textsc{metx} & \textsc{com} & \textsc{metx} \\ \hline \hline
       	NLLB-3B & 0.62 & 3.08 & 0.59 & 2.99 & 0.71 & 0.99 & 0.76 & 2.22 & 0.69 & 2.01 & 0.61 & 2.55 & 0.7 & 1.53 & 0.71 & 1.77 & 0.72 & 1.76 \\
	aya101-13B & 0.64 & 2.61 & 0.63 & 1.82 & 0.69 & 1.07 & 0.76 & 2.26 & 0.68 & 2.14 & 0.56 & 2.94 & 0.69 & 1.55 & 0.69 & 1.85 & 0.7 & 1.89 \\
	aya-e-8B & \textbf{0.7} & \textbf{1.9}& 0.65 & 1.81 & 0.72 & \textbf{0.85} & 0.77 & \textbf{2.01} & \textbf{0.71} & \textbf{1.82} & 0.53 & 3.02 & 0.71 & \textbf{1.3} & 0.72 & \textbf{1.47} & \textbf{0.73} & \textbf{1.59} \\
	babel-9B  & 0.64 & 3.06& 0.66 & 1.96 & 0.68 & 1.31 & 0.76 & 2.1 & 0.69 & 2.06 & 0.53 & 3.74 & 0.68 & 2.42 & 0.69 & 2.42 & 0.71 & 1.91 \\
	cohere-7B & 0.68 & 2.38 & 0.65 & 1.87 & 0.7 & 1.0 & 0.76 & 2.13 & 0.69 & 1.95 & 0.51 & 3.99 & 0.66 & 2.08 & 0.68 & 2.24 & 0.72 & 1.72 \\
	eurollm-9B & 0.7 & 1.99 & 0.66 & 1.65 & 0.72 & 0.89 & 0.77 & 2.13 & 0.7 & 1.87 & 0.6 & 2.65 & 0.45 & 6.53 & 0.72 & 1.57 & 0.72 & 1.67 \\
	madlad-10B & 0.65 & 2.67& 0.59 & 2.61 & 0.67 & 1.64 & 0.75 & 2.47 & 0.66 & 2.71 & 0.41 & 5.12 & 0.62 & 2.64 & 0.66 & 2.66 & 0.7 & 2.1\\
	madlad-3B  & 0.65 & 2.8 & 0.58 & 2.77 & 0.66 & 1.73 & 0.73 & 2.79 & 0.64 & 2.81 & 0.42 & 5.1 & 0.62 & 2.64 & 0.65 & 2.83 & 0.69 & 2.2\\
	mistral-7B & 0.32 & 8.73 & 0.55 & 3.12 & 0.62 & 1.85 & 0.73 & 2.7 & 0.62 & 2.96 & 0.3 & 8.42 & 0.52 & 4.52 & 0.61 & 3.38 & 0.65 & 2.89  \\
	qwen-7B & 0.53 & 4.74& 0.64 & 2.09 & 0.63 & 1.76 & 0.72 & 2.39 & 0.64 & 2.63 & 0.25 & 7.38 & 0.64 & 2.65 & 0.61 & 3.12 & 0.66 & 2.49  \\
	Llama3.1-8B & 0.54 & 4.68& 0.64 & 1.91 & 0.69 & 1.17 & 0.76 & 2.25 & 0.68 & 2.17 & 0.57 & 3.05 & 0.68 & 1.81 & 0.68 & 2.18 & 0.7 & 2.01  \\
	Llama3.2-3B & 0.43 & 6.84 & 0.55 & 3.11 & 0.63 & 1.76 & 0.73 & 2.62 & 0.62 & 2.86 & 0.49 & 4.5 & 0.62 & 2.7 & 0.54 & 5.27 & 0.66 & 2.66 \\
	Llama3.3-70B & 0.6 & 3.32 & \textbf{0.68} & \textbf{1.62} & \textbf{0.73} & 0.85 & 0.77 & \textbf{2.04} & \textbf{0.71} & 1.84 & \textbf{0.63} & \textbf{2.41} & \textbf{0.72} & 1.35 & 0.72 & 1.55 & 0.73 & 1.64 \\
	Tower-7B & 0.3 & 7.96 & 0.61 & 2.33 & 0.67 & 1.44 & 0.74 & 2.74 & 0.66 & 2.6 & 0.41 & 6.54 & 0.51 & 4.62 & 0.66 & 2.52 & 0.68 & 2.5  \\
    \hline

    \end{tabular}
    }
    \caption{Averaged results on CometKiwi (COM) and MetricX (etx) at the sentence-level from 9 \bouquet languages (top) and into (bottom). Best results are in bold (before rounding to 2 decimals). Best results on CometKiwi tend to be with Llama-3.3-70B (the largest model) and best results in MetricX tend to be with Aya-expanse-8B. Best direction is from and into English \label{tab:resultslanguages}.}
\end{table*}

\end{document}